\documentclass{midl} 

\usepackage{mwe} 

\usepackage[utf8]{inputenc} 
\usepackage[T1]{fontenc}    
\usepackage{hyperref}       
\usepackage{url}            
\usepackage{booktabs}       
\usepackage{amsfonts}       
\usepackage{nicefrac}       
\usepackage{microtype}      
\usepackage{graphicx}
\usepackage{float}
\usepackage{amsmath}
\usepackage{wrapfig,lipsum,booktabs}

\usepackage{array}
\usepackage{multirow}

\usepackage{titlesec}
\title[Leveraging Transfer Learning and MIL for H\&E HER2 Automatic Scoring]{Leveraging Transfer Learning and Multiple Instance Learning for HER2 Automatic Scoring of H\&E Whole Slide Images}
\midlauthor{\Name{Rawan S. Abdulsadig\nametag{$^{1}$}\midlotherjointauthor}
            \Email{r.abdulsadig@imperial.ac.uk}\\
            \Name{Bryan M. Williams\nametag{$^{1}$}\midlotherjointauthor}
            \Email{b.williams6@lancaster.ac.uk}\\
            \Name{Nikolay Burlutskiy\nametag{$^{2}$}\midljointauthortext{Corresponding author.}} \Email{nikolay.burlutskiy@astrazeneca.com}\\
            \addr $^{1}$Lancaster University, UK\\
            \addr $^{2}$AstraZeneca Cambridge, UK \\
}

\begin{document}

\maketitle

\begin{abstract}
    Expression of human epidermal growth factor receptor 2 (HER2) is an important biomarker in breast cancer patients who can benefit from cost-effective automatic Hematoxylin and Eosin (H\&E) HER2 scoring. However, developing such scoring models requires large pixel-level annotated datasets. Transfer learning allows prior knowledge from different datasets to be reused while multiple-instance learning (MIL) allows the lack of detailed annotations to be mitigated. The aim of this work is to examine the potential of transfer learning on the performance of deep learning models pre-trained on (i) Immunohistochemistry (IHC) images, (ii) H\&E images and (iii) non-medical images. A MIL framework with an attention mechanism is developed using pre-trained models as patch-embedding models. It was found that embedding models pre-trained on H\&E images consistently outperformed the others, resulting in an average AUC-ROC value of $0.622$ across the 4 HER2 scores ($0.59-0.80$ per HER2 score). Furthermore, it was found that using multiple-instance learning with an attention layer not only allows for good classification results to be achieved, but it can also help with producing visual indication of HER2-positive areas in the H\&E slide image by utilising the patch-wise attention weights.
\end{abstract}


\section{Introduction and Related Work}\label{relatedWork}

When suspecting breast cancer, pathologists perform microscopic analysis of breast tissue samples obtained using a needle biopsy \cite{american2019breast}. The sample is stained with Hematoxylin and Eosin (H\&E) to examine the presence of cancer and with Immunohistochemistry (IHC) for the assessment of human epidermal growth factor receptor 2 (HER2), which helps to indicate whether the patient is eligible for targeted anti-HER2 therapies. There is an interest in developing automatic deep learning approaches for IHC HER2 scoring to reduce pathologists' workload \cite{qaiser2018her}. An automatic H\&E HER2 scoring is even more preferable and cost effective approach than using IHC HER2 staining \cite{oliveira2020weakly}.

Model-based transfer learning is a common technique in deep learning where the knowledge learnt from a source-domain is "transferred" to a different domain, then adapted to suit the problem domain of concern. This technique helps cut down the computational time needed and most importantly the target-domain data required to train the model \cite{zhuang2019comprehensive}. Transfer learning becomes especially handy when dealing with medical data that is either scarce or hard to obtain, more specifically in the case of medical imaging data. 

To date, few studies have attempted to pre-train models with histopathology images due to a lack of labeled images and experimental studies comparing the effect of transfer learning from different medical imaging domains. \citeauthor{sun2017comparison} evaluated the effect of transfer learning from ImageNet to H\&E images. The transferred model captured Gabor filters derived and colour specific features generalisable across different types of histopathology images.

Multiple-instance learning (MIL) has been shown to be efficient when there is lack of detailed annotations since the method is based on the assumption that negatively labeled bags of patches only contain negative instances, while positive bags contain at least one positive patch \cite{carbonneau2018multiple} mitigating the need for accurate annotations. \citeauthor{courtiol2018classification} used an ImageNet pre-trained ResNet model for slide-level classification of TCGA-Lung and Camelyon-16 datasets. \citeauthor{oliveira2020weakly} attempted to use IHC patches to pre-train an embedding model in a MIL framework for predicting HER2 scores from H\&E whole slide images (WSIs). Although the paper showed that using IHC patches for pre-training the model helped to achieve high accuracy, the study did not show the performance of pre-training on different source domains.

\section{Methodology}\label{methods}


Three types of source classification tasks were examined:
1. Different type of images and a different classification task: a dataset of non-medical images. 2. Same type of images but a different classification task: H\&E images of a different type of tissue and annotated with different information. 3. Different type of images but the same classification task: IHC images of breast tissue which show the information of concern, and annotated with HER2.\\

\begin{figure}[ht]
\centering
  \includegraphics[width=14cm]{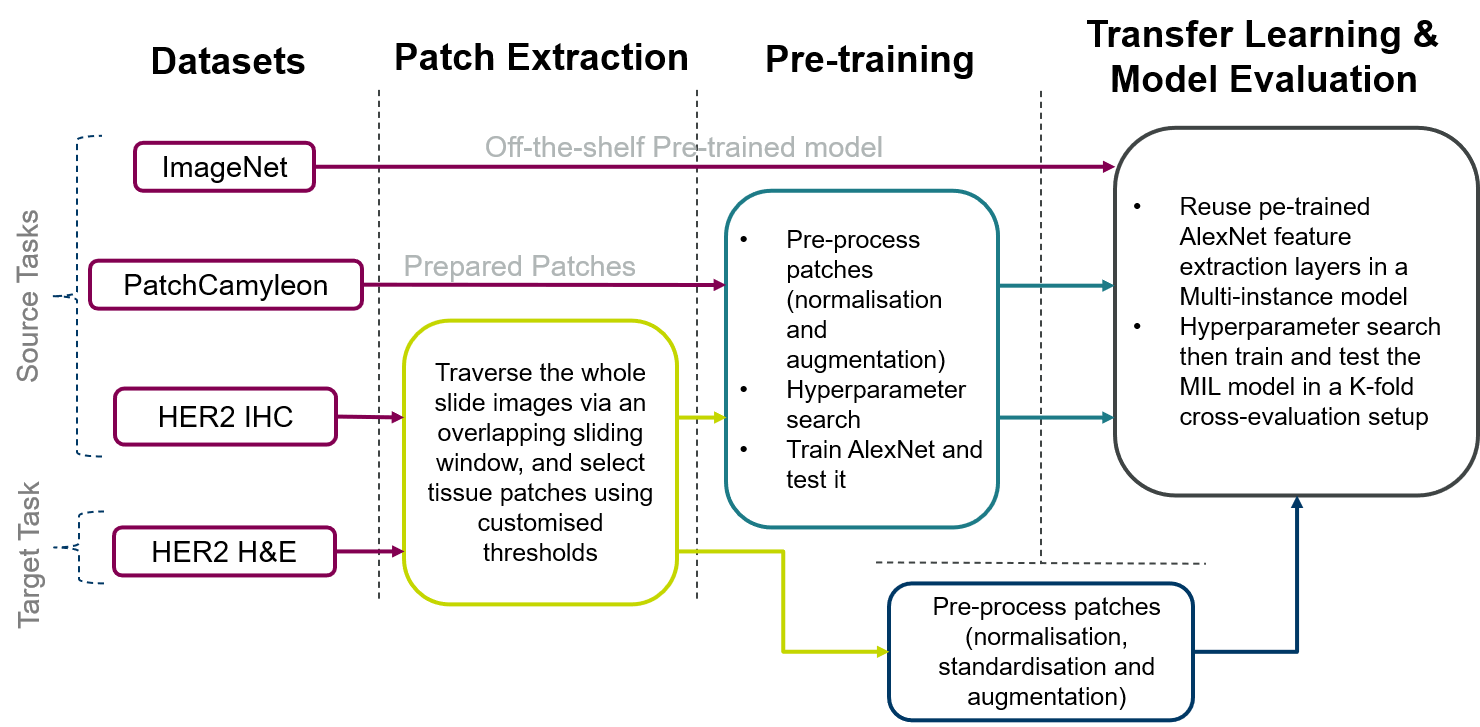}
\caption{An overview of the methodology of the work.}
\label{methodologyOverview}
\end{figure}

The \href{https://warwick.ac.uk/fac/sci/dcs/research/tia/her2contest/}{HER2 Scoring Contest} dataset of H\&E and IHC whole slide images was considered to be the main dataset providing the H\&E stained WSIs annotated with their HER2 scores, in addition to the IHC stained WSIs representing the third type of source task (i.e. different type of images, same classification task). The \href{https://github.com/basveeling/pcam}{PatchCamelyon} dataset was chosen to represent the second type of source task (i.e. same type of images, different classification task) as it was found to be well prepared for training in terms of the annotations, the balanced classes and the size of the images. On the other hand, \href{http://www.image-net.org/}{ImageNet} was chosen to represent the first source task of non-medical images (i.e. different type of images, different classification task) since it is a rich dataset that is often used to pre-train image classification models in practice. Figure \ref{methodologyOverview} illustrates a general overview of the methodology\footnote{The source code for the implemented pipleine is available via a GitHub \href{https://github.com/RawanSaifAldeen/HER2-Scoring_MIL-Attention-Model}{repository}}.
\\

Patches were extracted from the H\&E and IHC WSIs at 0.5 $\mu$m magnification level using handcrafted thresholds operating on the hue, saturation and value (HSV) colour space. The patches were augmented by randomly varying brightness, contrast, saturation and hue, as well as applying affine transformations (rotations, shear, translation and scaling), in addition to Gaussian noise.

\begin{figure}[H]
\centering
  \includegraphics[width=13 cm ]{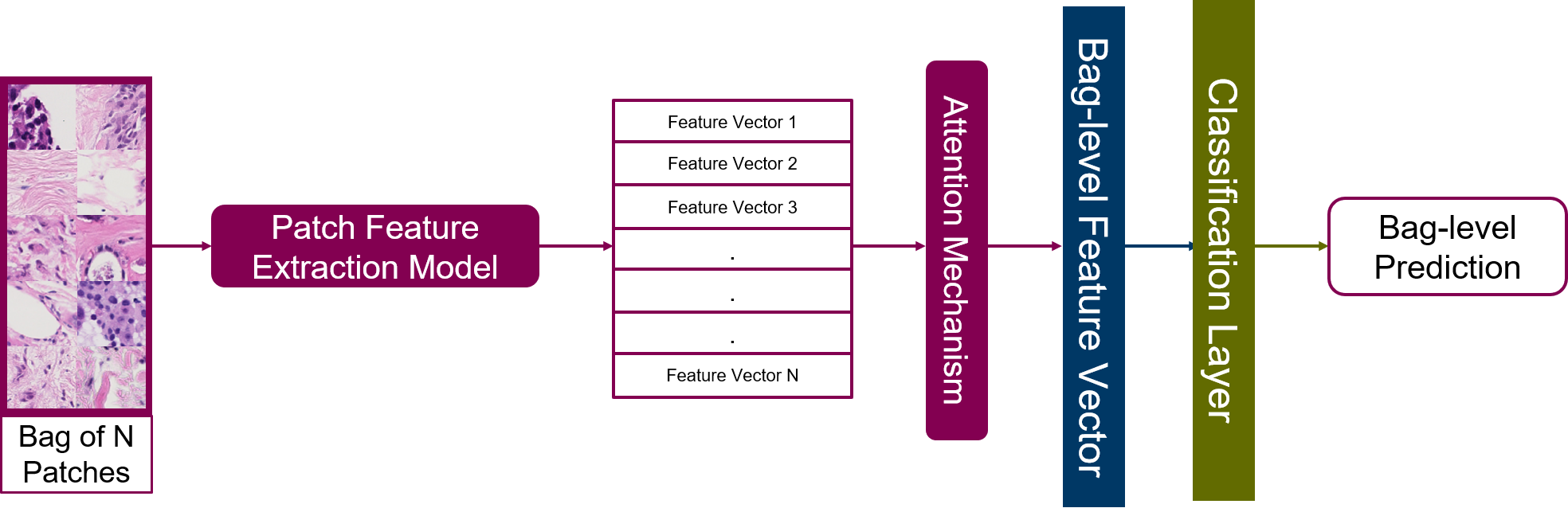}
\caption{The implemented multiple-instance learning approach}
\label{MILDiagram}
\end{figure}

Figure \ref{MILDiagram} shows the implemented MIL approach which heavily relays on prior knowledge obtained by the transferred patch feature-extraction model pre-trained on a certain source task.

Referring to Figure \ref{MILDiagram} above, when a set of $N$ patches was uniformly extracted from a WSI to form a bag, each patch in that bag was transformed into a 1-dimensional feature vector using the pre-trained convolutional layers of a pre-trained AlexNet, that set of layers was then treated as the patch feature-extraction model.

The resulting $N$ feature vectors were then fed to an attention layer, the output of the attention layer was a weighted sum of the $N$ feature vectors representing the bag, this operated as a pooling operation that ensured shift-invariance \cite{ilse2018attention}. The output of the attention mechanism represented a bag-level feature vector which was then fed to the classification layer.\\

The attention weights were learnt during training on the target task, an attention weight $a_k$ associated with feature vector $\boldsymbol{V}_k$ (where $k = 1,2,3,...,N$) is defined as:

\begin{equation}
    a_k = \text{Softmax}(\boldsymbol{w}_1^{T}\text{Tanh}(\boldsymbol{w}_2\boldsymbol{V}_k^{T}))
\end{equation}

where $\boldsymbol{w}_1$ and $\boldsymbol{w}_2$ are learnt weights of two linear layers. The output of the attention layer $\boldsymbol{A}$ is then:

\begin{equation}
    \boldsymbol{A} = \sum_{k=1}^K a_k\boldsymbol{V}_k
\end{equation}
Finally, the attention weights were utilised to infer the contribution of each patch in the bag to predict a specific HER2 score.

A further utilisation of the attention weights of a trained MIL model was to infer the contribution of each patch in the bag offered to the model for it to predict a specific HER2 score, and thereafter to visualize the locations of those patches in an overlaid heatmap.\\

\begin{wrapfigure}{r}{0.5\linewidth}
\centering
  \includegraphics[width=8 cm ]{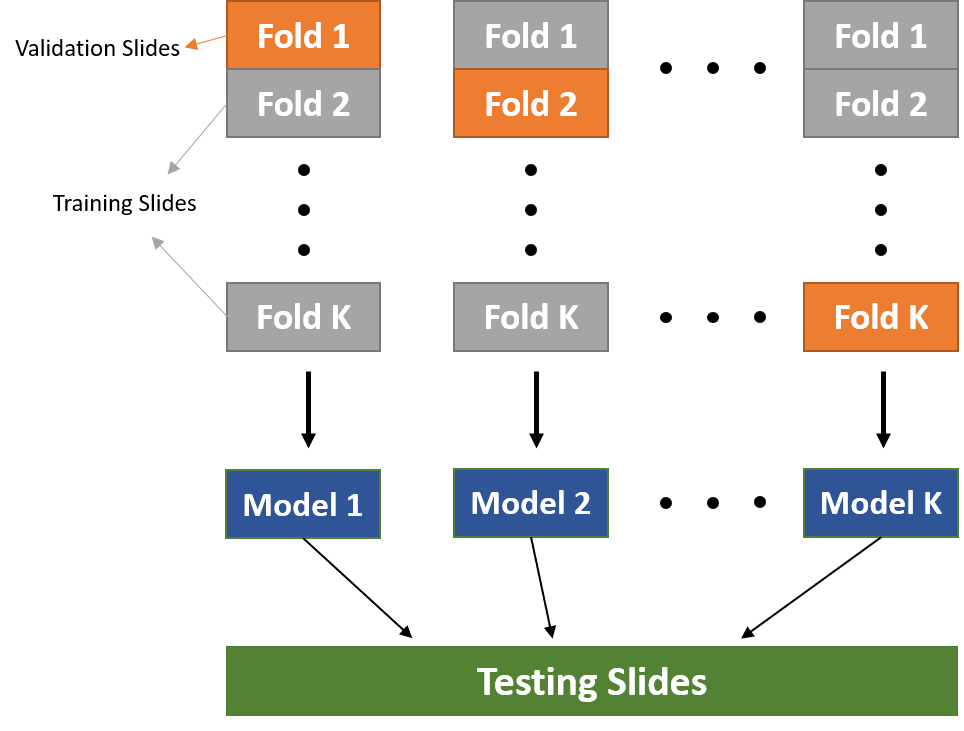}
\caption{The K-fold cross-evaluation approach}
\label{kfoldCVdiagram}
\end{wrapfigure}

Evaluating the variation in the classification performance of each model type was done by splitting the training set into K partitions or ''folds'' where each model is trained on a distinct set of K-1 folds and validated on the remaining fold. This was done in order to obtain K variations of the model type representing models that each learnt from different sections of the data. The K trained models were then tested on a fixed set of testing slides and the evaluation metrics were then calculated for each of the K models when tested on that set. Figure \ref{kfoldCVdiagram} shows a diagram of the approach used for evaluating the models.\\

Quantifying the variation in the performance of the models was done by calculating the confidence interval around a mean value of the evaluation metrics achieved by the K models, a 95\% confidence interval was calculated as follows:
\begin{equation}
    \text{CI (95\%)}  = \text{Mean} \pm 1.96*\text{Standard Error} = \text{Mean} \pm 1.96*\frac{\text{Standard Deviation}}{\sqrt{K}}
\end{equation}
In this calculation, the values of the metrics obtained when testing on each of the K folds were used as an approximation to their expected values at that fold.

\section{Experiments and Results}\label{results}

Obtaining prior knowledge from the three source tasks was done by pre-training an AlexNet CNN architecture on ImageNet, PatchCamelyon and the constructed IHC HER2 (52 WSIs providing $\approx 380,000$ $224\times 224$ patches) datasets. For ImageNet, PyTorch's off-the-shelf pre-trained AlexNet model was used, while pre-training on PatchCamelyon (PCAM) and IHC was carried out locally and tested to obtain their performance measures in their respective classification tasks. Table \ref{pretrainingTable} shows the average accuracy and the AUC-ROC value obtained in each case. The convolutional layers of these pre-trained AlexNet models were then used in the MIL framework as patch-embedding models.

\begin{table}[ht!]
\centering
\begin{tabular}{|m{3cm}|ccc|c|}
\hline
\textbf{Source Task} & \textbf{Classes} & \textbf{Accuracy} & \textbf{AUC-ROC} \\
\hline
ImageNet (2012) & 1000  & 60\% & - \\
PCAM & 2  & 77\% & 0.89 \\
IHC HER2 & 4  & 40\% & 0.68\\
\hline
\end{tabular}
\caption{The performance of the resulting AlexNet models on each of the three source tasks}
\label{pretrainingTable}
\end{table} 

Considering the multi-class classification problem of HER2 scoring, bags of 100 patches were formed by sampling eligible patches from the selected slides during the 5-fold cross evaluation training. The slides were split into 44 training slides and 8 testing slides. Each epoch was done by training the MIL model on 6,400 bags of patches and validated on 2,500 bags, and those models were then tested on 2,500 bags from the testing slides.  Those bags were partitioned in batches of 64 bags.

The training bags were not repeated during the training process, every epoch introduced different sampled bags. This was done as a bag-level augmentation method in order to cover all possible variations of bags that can be constructed from each training slide, while validation and testing bags were kept fixed.\\

The weights of the patch-embedding models remained frozen during the training of the MIL model, while the rest of the layers were updated. The choice to freeze the patch-embedding model weights was made to avoid complexity of training and to explore the effect of the un-modified prior knowledge these models provided on the classification performance.

Adam optimiser was used for training, and the optimisation parameters were searched for in a grid of learning rate values [1e-03,1e-04, 1e-05, 1e-06] against weight decay values [1e-03,1e-04,1e-05,1e-06, 1e-07]. The set of values resulting in the minimum validation loss were used to train each model type in the 5-fold cross-evaluation setup, these models were then tested and their performance metrics were obtained along with their 95\% confidence intervals.\\

Table \ref{resultsTable} shows the average precision, recall, F1-score and AUC-ROC values across the 4 HER2 classes and their 95\% confidence intervals when evaluating the different types of MIL models using the 5-fold cross-evaluation approach. The MIL models with PatchCamyleon (PCAM) pre-trained patch-embedding model seemed to outperform the other types in all evaluation measures, suggesting that the provided patch-embeddings were capturing most of the useful information for more accurate overall classification.\\

\begin{table}[H]
\centering
\begin{tabular}{|m{3.15cm}|cccc|}
\hline
\textbf{Patch-embedder Pre-trained on:} & \textbf{Precision} & \textbf{Recall} & \textbf{F1-Score} & \textbf{AUC-ROC}\\
\hline
None (Random) & $0.103 \pm 0.052$ & $0.245 \pm 0.142$ & $0.127 \pm 0.056$ & $0.471 \pm 0.054$\\
\textbf{PCAM} & $\boldsymbol{0.370 \pm 0.049}$ & $\boldsymbol{0.369 \pm 0.051}$ & $\boldsymbol{0.327 \pm 0.058}$ & $\boldsymbol{0.622 \pm 0.027}$\\
IHC & $0.349 \pm 0.176$ & $0.247 \pm 0.155$ & $0.213 \pm 0.136$ & $0.511 \pm 0.136$\\
ImageNet & $0.177 \pm 0.041$ & $0.189 \pm 0.050$ & $0.169 \pm 0.040$ & $0.454 \pm 0.112$\\\hline
\end{tabular}
\caption{Average precision, recall, F1-score and AUC-ROC values across the 4 classes and their 95\% confidence bounds}
\label{resultsTable}
\end{table}

\begin{wrapfigure}{r}{0.5\linewidth}
\centering
\includegraphics[width=7.7cm]{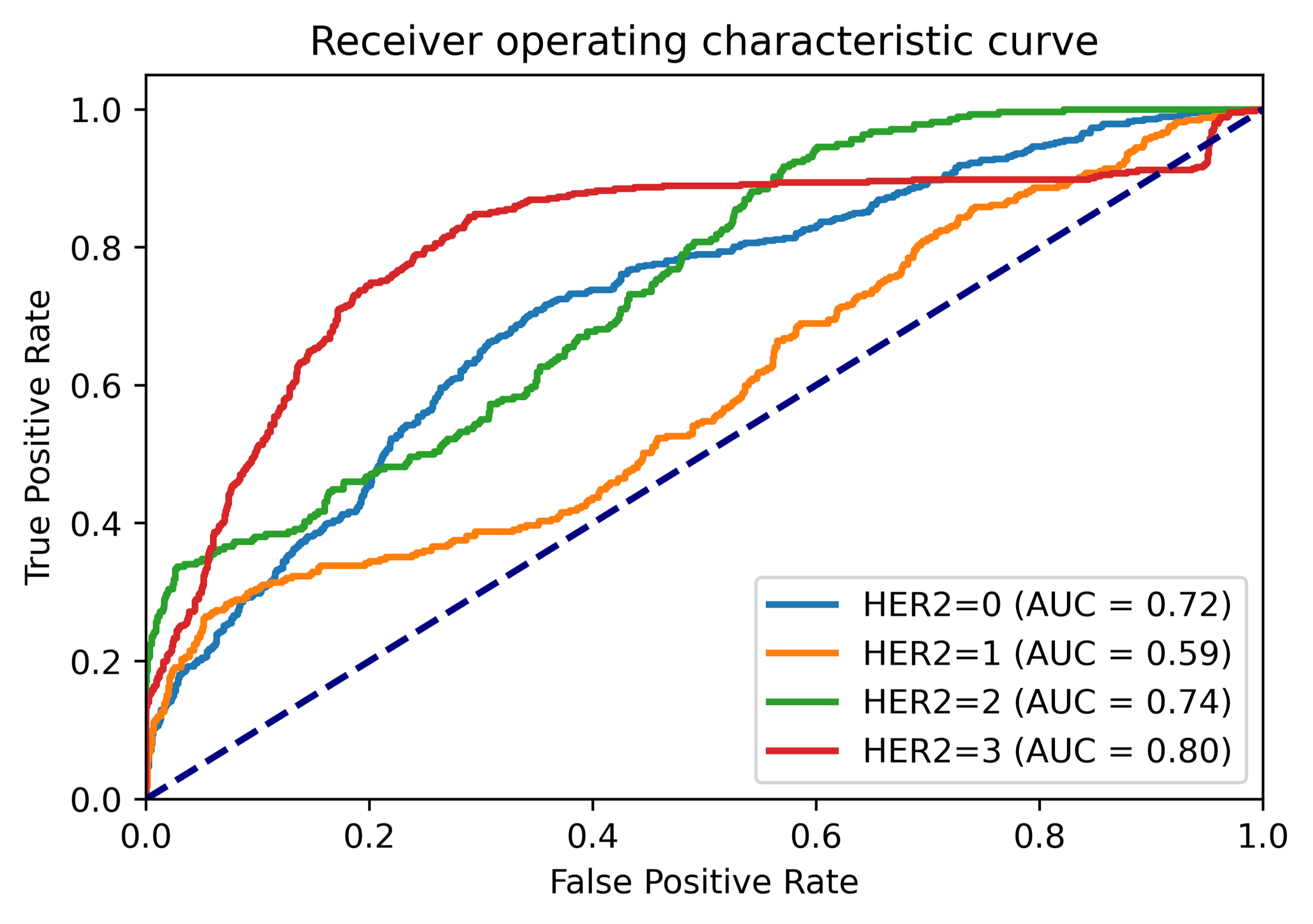}
\caption{ROC curve for the best MIL model pre-trained on PCAM.}
\label{ROC's}
\end{wrapfigure}

Figure \ref{ROC's} shows the receiver operator characteristic curves and AUC-ROC values corresponding to the 4 HER2 classes when testing one of the best performing MIL models equipped with PCAM pre-trained patch-embedding model.
Table \ref{resultsTable} and figure \ref{ROC's} suggest that PatchCamyleon (PCAM) was the better source domain for providing good prior knowledge that resulted in an acceptable classification performance across the 4 classes.\\

Having a well trained MIL model that predicts the HER2 scores at bag-level, inferring the slide-level HER2 score of an unseen slide can be done by taking a ''Monte-Carlo'' type of approach where bags of patches are repeatedly sampled with replacement, then fed to the MIL model for prediction in each ''simulation''. Those predictions can then be aggregated to obtain the expected overall slide-level HER2 score. Given a good performing model, increasing the number of sampled bags (or ''simulations'') is expected to increase the accuracy and the certainty of the final prediction.\\

During the process of predicting the slide-level HER2 score, the attention mechanism can be utilised during the repeated sampling which allows to obtain an estimation of the positivity of the individual patches. One way to achieve that is by weighting the predicted bag-level HER2 scores with the value of the attention weight corresponding to each patch, then aggregating those weighted predictions to obtain the patch-level HER2 grading. Algorithm \ref{heatmapsudocode} illustrates the general approach to get those patch-level probabilities of being HER2 positive.

\begin{algorithm}[H]
\SetKwInput{KwInput}{Input}
 \KwInput{Whole slide image (WSI)}
 \KwResult{Patch locations and their corresponding probabilities of being HER2 positive}
 \For{$i\gets1$ \KwTo $N\_samples$}{
  bag, locations $\gets$ Get\_Bag(WSI)\\
  bag\_probability, attention\_weights $\gets$ MIL\_Predict(bag)\\
  patch\_probabilities $\gets$ bag\_probability $\times$ attention\_weights\\
  Update\_Heatmap(locations, patch\_probabilities)\\
 }
 \caption{Patch-level prediction using an MIL model}
  \label{heatmapsudocode}
\end{algorithm}

\begin{wrapfigure}{r}{0.6\linewidth - 10 cm}
\centering
  \includegraphics[width=10 cm ]{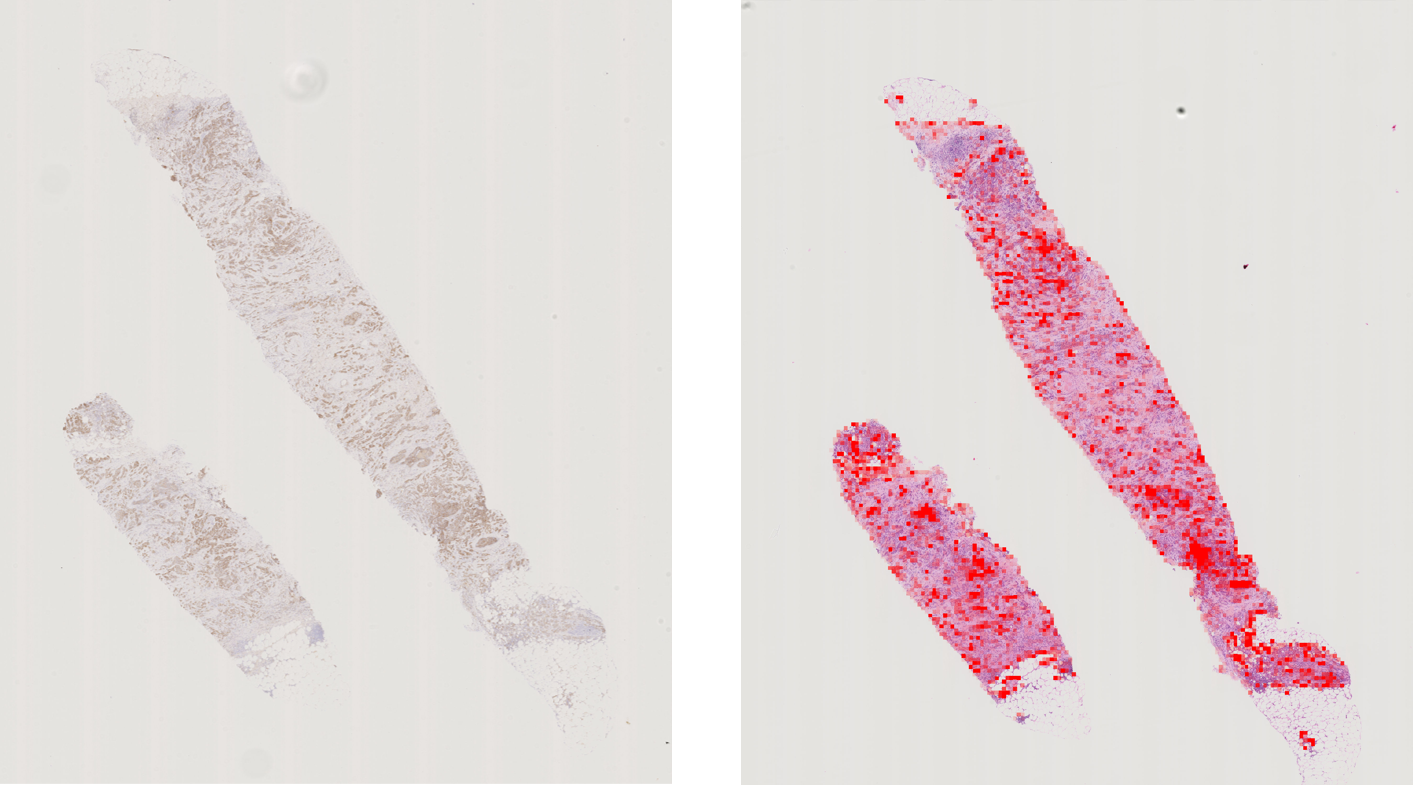}
\caption{IHC WSI (left) and its corresponding H\&E WSI of the same tissue with an overlaid heatmap of patch-level probabilities of being HER2 positive}
\label{heatmapfigure}
\end{wrapfigure}

Following the approach described above, a sample WSI was used to produce a heatmap of patch-level probabilities of being HER2 positive (score 2 or 3). The best performing MIL model with PCAM pre-trained patch-embedding model was used to produce the attention weights and the bag-level predicted probabilities of being positive. The Update\_Heatmap function was implemented to append the probabilities to lists that correspond to the location of the patches in the slide in order to take the mean value of the probabilities assigned to each patch as a final estimation. Figure \ref{heatmapfigure} shows the resulting heatmap when 1000 bags where sampled from the H\&E slide and its corresponding IHC slide.

\section{Conclusions and Future Work}\label{conclusion}

A cost-effective automatic H\&E HER2 scoring model was developed. Transfer learning from a similar staining, H\&E to H\&E, demonstrated to be more beneficial than from IHC to H\&E images neither than from non-medical ImageNet to H\&E images. Potentially, a further hyper-parameter tuning could improve IHC to H\&E transfer which was suggested by the wide confidence intervals. In this paper, MIL demonstrated to be a promising approach where only a limited number of slide level annotated WSIs were available. Optimising the models and exploring fine-tuning strategies for the MIL pipeline could further boost performance and provide a better understanding of transfer learning contribution across different datasets.


\bibliography{HER2Paper2021} 
\end{document}